\documentclass[runningheads]{llncs}
\usepackage[T1]{fontenc}
\usepackage{hyperref}
\usepackage{amsmath}
\usepackage{xcolor}
\usepackage{multirow}

\usepackage{graphicx}
\begin{document}
\title{Fusion of Foundation and Vision Transformer Model Features for Dermatoscopic Image Classification
}
%
%
\author{Amirreza Mahbod\inst{1} \and
Rupert Ecker\inst{2} \and
Ramona Woitek\inst{1}
}
\authorrunning{A. Mahbod et al.}
%
\institute{Research Center for Medical Image Analysis and Artificial Intelligence, Department of Medicine, Faculty of Medicine and Dentistry, Danube Private University, Austria 
\and
Department of Research and Development, TissueGnostics GmbH, Austria
}
\maketitle              

{\begin{abstract}
		
Accurate classification of skin lesions from dermatoscopic images is essential for diagnosis and treatment of skin cancer. In this study, we investigate the utility of a dermatology-specific foundation model, PanDerm, in comparison with two Vision Transformer (ViT) architectures (ViT base and Swin Transformer V2 base) for the task of skin lesion classification. Using frozen features extracted from PanDerm, we apply non-linear probing with three different classifiers, namely, multi-layer perceptron (MLP), XGBoost, and TabNet. For the ViT-based models, we perform full fine-tuning to optimize classification performance. Our experiments on the HAM10000 and MSKCC datasets demonstrate that the PanDerm-based MLP model performs comparably to the fine-tuned Swin transformer model, while fusion of PanDerm and Swin Transformer predictions leads to further performance improvements. Future work will explore additional foundation models, fine-tuning strategies, and advanced fusion techniques.

\end{abstract}}
\section{Background}\label{sec1}
Skin cancer is one of the most widespread types of cancer, with a rising incidence worldwide~\cite{torbati2025multi}. Although specific types of skin cancer such as melanoma can be life-threatening, early detection significantly increases patient survival rates~\cite{10.1001/jamadermatol.2021.3884}. While histological image analysis remains the gold standard for diagnosis and grading, dermatoscopic image analysis is more suitable for early screening, as it is non-invasive and cost-effective compared to histopathology~\cite{torbati2025multi,Elmorej2813}.

To enhance and accelerate dermatoscopic image analysis, numerous computerized methods have been proposed in the literature~\cite{10.3389/fmed.2023.1305954}. Early approaches typically involved skin lesion segmentation, handcrafted feature extraction and selection, followed by training a classifier~\cite{MAHBOD2020105725}. In contrast, recent studies rely heavily on deep learning, particularly automatic feature extraction using pre-trained convolutional neural networks (CNNs) and vision transformers (ViTs)~\cite{9412307,Ayas2023}.

Popular CNN-based models include ResNet, VGG, and EfficientNet~\cite{9412307,MAHBOD2020105475,mahbod2019skin}, while ViT-based architectures such as the Vision Transformer and Swin Transformer~\cite{Ayas2023,Adebiyi2024.09.19.24314004} have also been applied to the task of skin lesion image classification.

More recently, there has been a significant shift toward developing and utilizing large-scale pre-trained foundation models for a wide range of computer vision tasks, both medical and non-medical. Several domain-specific foundation models have been introduced—such as UNI for histopathology~\cite{Chen2024} and others for radiological imaging~\cite{Pai2024}. In the dermatology domain, a new model called PanDerm was recently proposed, serving as a foundation model specifically trained for skin lesion image analysis~\cite{yan2024general}.

Although many foundation models have demonstrated impressive performance in zero-shot applications for various downstream image analysis tasks, their performance on task-specific applications, particularly when compared to CNN or ViT-based models has not been thoroughly investigated. For instance, a recent study by Ganz et al.~\cite{10.1007/978-3-658-47422-5_15} on mitotic figure classification in histological images evaluated the performance of several foundation models against standard CNN and ViT feature extractors, as well as end-to-end CNN training. Interestingly, their results showed superior performance for end-to-end CNN training compared to linear probing with foundation model features.

In this study, we aim to evaluate the classification performance of PanDerm, a foundation model trained specifically for skin lesion analysis, using non-linear probing, and compare it with two standard ViT-based models for the dermatoscopic image classification task. Furthermore, we explore whether combining predictions from PanDerm and ViT models can lead to improved classification performance.

\section{Method}\label{sec1}
As baseline classification models, we utilize the ViT base and Swin Transformer V2 base architectures, both pre-trained on the ImageNet dataset~\cite{Deng2009}. The classification heads of both networks are removed and replaced with custom multilayer perceptrons (MLPs). The ViT model outputs a 768-dimensional feature vector, while the Swin Transformer model produces a 1024-dimensional feature vector. These feature vectors are passed through MLPs consisting of a hidden layer with ReLU activation, a dropout layer, and a final linear layer that maps to the number of skin lesion classes. We fine-tune the entire model, rather than training only the added MLP layers, as full fine-tuning has been shown to yield better performance in other image classification tasks~\cite{10.1007/978-3-658-47422-5_15}. For training, all images are resized to 224$\times$224 pixels and normalized using the mean and standard deviation of the ImageNet dataset. A set of standard data augmentation techniques, including random rotation, horizontal and vertical flipping, and color jittering, is applied during training to improve model robustness. 

For classification using the PanDerm foundation model, we first extract image embeddings using the frozen vision encoder of the pre-trained model. These extracted features are then used to train a set of well-known classifiers~\cite{Dwivedi2025.05.02.25326847}. In this study, we apply non-linear probing using three different models, namely MLP, XGBoost~\cite{10.1145/2939672.2939785}, and TabNet~\cite{Arik_Pfister_2021}.

\begin{figure}[h]
	\centering
	\includegraphics[width=0.6\textwidth]{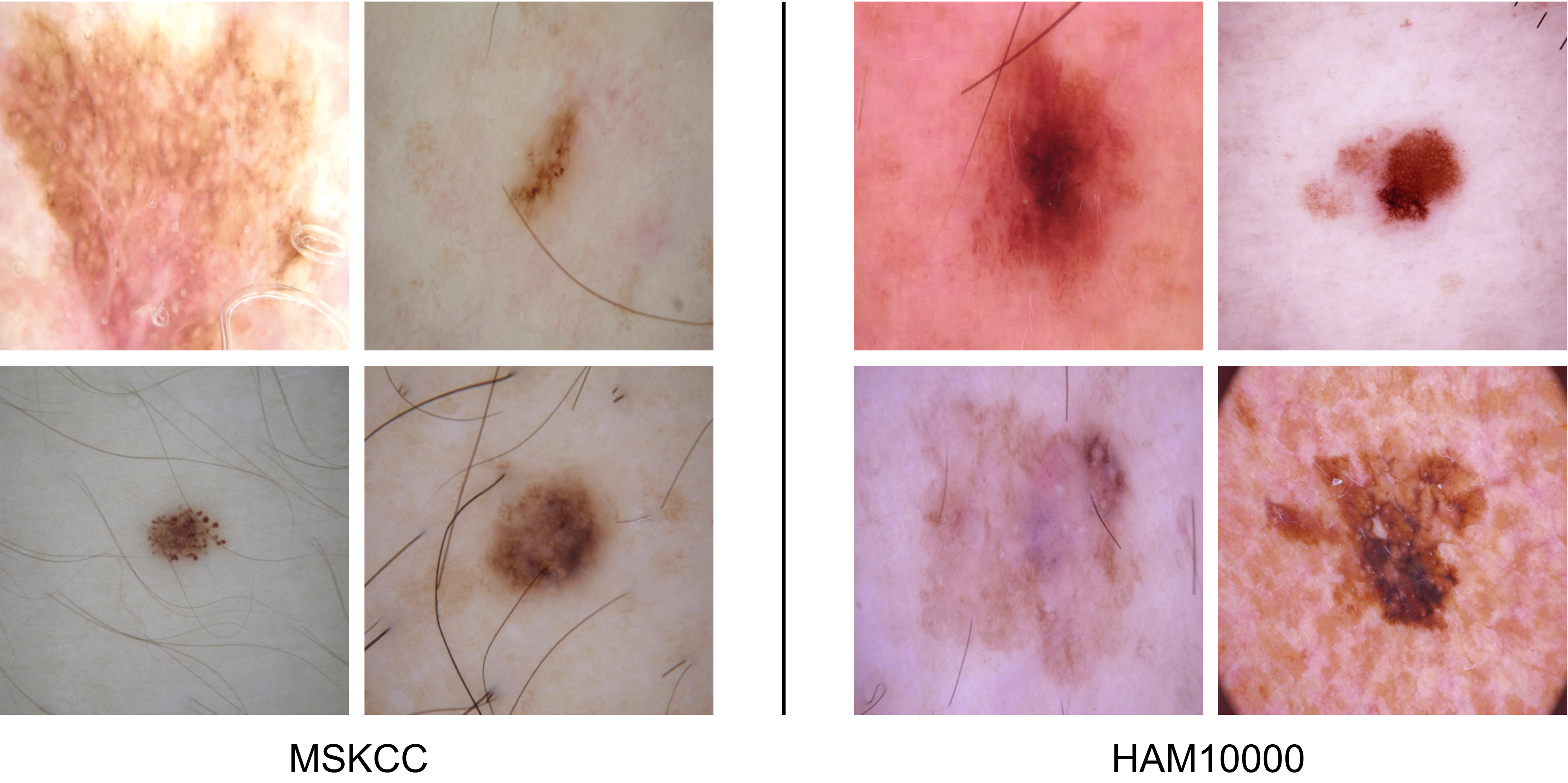}
	\caption{Example images from the MSKCC and HAM10000 datasets.}\label{fig:examples}
\end{figure}

To conduct our experiments, we use two datasets: MSKCC~\cite{Codella2017} and HAM10000~\cite{tschandl2018ham10000}. The MSKCC dataset contains 8,984 dermatoscopic images labeled into two classes: melanoma and others. The HAM10000 dataset includes 10,015 dermatoscopic images across seven skin lesion categories. We follow the official train, validation, and test splits provided in~\cite{yan2024general}. Specifically, for the MSKCC dataset, we use 6,189 images for training, 1,131 for validation, and 1,664 for testing. For the HAM10000 dataset, we use 8,207 for training, 575 for validation, and 1,232 for testing. 
Example images from each dataset are depicted in Figure~\ref{fig:examples}.
To evaluate the classification performance, we use accuracy and balanced accuracy, both calculated on the test set for each dataset.

\section{Preliminary Results and Discussion}
The main results from our experiments are presented in Table~\ref{tab:res}. In addition to reporting the performance of the baseline ViT-based classification models and PanDerm features with different classifiers, we also include the results of a fusion-based approach, shown in the last row of the table. This approach combines predictions from the best-performing baseline ViT model (Swin Transformer V2 base) and the best-performing PanDerm-based model (MLP trained on PanDerm embeddings).

The reported results indicate that among the classifiers applied to PanDerm features, the MLP classifier consistently delivers superior performance compared to XGBoost and TabNet across both datasets. Moreover, the baseline Swin Transformer V2 model achieves performance that is comparable to the PanDerm-based MLP model. Notably, the fusion-based approach achieves the best performance on the HAM10000 dataset and yields better or competitive results on the MSKCC dataset when compared to individual models. 

\begin{table}[]
	\centering

	\begin{tabular}{|lcc|lcc|}
		\hline
		\multicolumn{3}{|c|}{\textbf{MSKCC}}                                                                                                                                                             & \multicolumn{3}{c|}{\textbf{HAM10000}}                                                                                                                                                          \\ \hline
		\multicolumn{1}{|l|}{\textbf{Model}}   & \multicolumn{1}{c|}{\textbf{\begin{tabular}[c]{@{}c@{}}Acc. \\ (\%)\end{tabular}}} & \textbf{\begin{tabular}[c]{@{}c@{}}Bal. Acc. \\ (\%)\end{tabular}} & \multicolumn{1}{c|}{\textbf{Model}}   & \multicolumn{1}{c|}{\textbf{\begin{tabular}[c]{@{}c@{}}Acc. \\ (\%)\end{tabular}}} & \textbf{\begin{tabular}[c]{@{}c@{}}Bal. Acc. \\ (\%)\end{tabular}} \\ \hline
		\multicolumn{1}{|l|}{ViT}              & \multicolumn{1}{c|}{69.47}                                                         & 57.90                                                               & \multicolumn{1}{l|}{ViT}              & \multicolumn{1}{c|}{88.88}                                                         & 73.63                                                              \\ \hline
		\multicolumn{1}{|l|}{Swin}             & \multicolumn{1}{c|}{75.30}                                                          & \textbf{64.52}                                                              & \multicolumn{1}{l|}{Swin}             & \multicolumn{1}{c|}{91.80}                                                          & \underline{81.43}                                                              \\ \hline
		\multicolumn{1}{|l|}{PanDerm\_MLP}     & \multicolumn{1}{c|}{\underline{76.32}}                                                         & \underline{64.40}                                                              & \multicolumn{1}{l|}{PanDerm\_MLP}     & \multicolumn{1}{c|}{\underline{92.69}}                                                        & 79.57                                                              \\ \hline
		\multicolumn{1}{|l|}{PanDerm\_XGBoost} & \multicolumn{1}{c|}{75.60}                                                          & 63.39                                                              & \multicolumn{1}{l|}{PanDerm\_XGBoost} & \multicolumn{1}{c|}{89.93}                                                         & 59.84                                                              \\ \hline
		\multicolumn{1}{|l|}{Panderm\_TabNet}  & \multicolumn{1}{c|}{71.87}                                                         & 58.61                                                              & \multicolumn{1}{l|}{Panderm\_TabNet}  & \multicolumn{1}{c|}{91.15}                                                         & 72.96                                                              \\ \hline
		\multicolumn{1}{|l|}{Fusion}           & \multicolumn{1}{c|}{\textbf{76.80}}                                                          & 64.36                                                              & \multicolumn{1}{l|}{Fusion}           & \multicolumn{1}{c|}{\textbf{93.59}}                                                         & \textbf{83.25}                                                              \\ \hline
	\end{tabular}
		\caption{Classification performance of the baseline ViT-based models and the PanDerm-based models on the MSKCC and HAM10000 datasets.
			}
			\label{tab:res}
\end{table}

\section{Conclusion and Outlook}
In this work, we compare two standard ViT-based models with non-linear probing approaches that utilizes frozen features from a foundation model for the task of skin lesion image classification. Our results indicate that the ViT-based models deliver comparable performance to the foundation model features with non-linear classifiers. Furthermore, the classification performance can be improved through feature fusion. In future work, we plan to extend our study by incorporating additional models, including CNN-based, ViT-based, and both medical and non-medical foundation models, for a broader comparison. We also aim to investigate the impact of fine-tuning foundation models, as opposed to using frozen features, on classification performance. Additionally, we will explore alternative fusion strategies, such as early fusion, to evaluate their effect on the classification performance.

\section{Acknowledgment}
This project was partially supported from Horizon—the Framework Programme for Research and Innovation (2022–2027), Marie Sklodowska-Curie Actions of the European Union under Research Executive Agency grant agreement eRaDicate No. 101119427. 

%
%
\bibliographystyle{splncs04}
\bibliography{refs}

\end{document}